\documentclass{article}
\usepackage{spconf,amsmath,graphicx}
\usepackage{subcaption}
\usepackage{amsfonts}
\usepackage{mathtools}
\usepackage{bm}
\usepackage{hyperref}
\usepackage{booktabs}
\usepackage{tabularx}

\usepackage{multirow}
\title{Unsupervised Deep Clustering for Source Separation: \\ Direct Learning from Mixtures using Spatial Information }
\name{Efthymios Tzinis$^{\sharp}$ \qquad \thanks{Code:
\href{https://github.com/etzinis/unsupervised_spatial_dc}{github.com/etzinis/unsupervised\_spatial\_dc}}
      Shrikant Venkataramani$^{\sharp}$ \qquad 
      Paris Smaragdis$^{ \sharp \flat}$\thanks{Supported by NSF grant \#1453104}}
\address{$^{\sharp}$University of Illinois at Urbana-Champaign, Department of Computer Science \\ 
         $^{\flat}$Adobe Research}
\begin{document}
\ninept
\maketitle
\begin{abstract}
We present a monophonic source separation system that is trained by only observing mixtures with no ground truth separation information. We use a deep clustering approach which trains on multi-channel mixtures and learns to project spectrogram bins to source clusters that correlate with various spatial features. We show that using such a training process we can obtain separation performance that is as good as making use of ground truth separation information. Once trained, this system is capable of performing sound separation on monophonic inputs, despite having learned how to do so using multi-channel recordings.
\end{abstract}
\section{Introduction}
\label{sec:intro}
A central problem when designing source separation systems is that of defining what constitutes a source. Multi-microphone systems like \cite{smaragdis1998blind,parra2002geometric,saruwatari2003blind,sawada2006blind} make use of spatial information to define and isolate sources, component-based systems like \cite{smaragdis2007convolutive,virtanen2007monaural}, use pre-learned dictionaries to identify sources, and more recently, neural net-based models \cite{huang2014deep,wang2018multi} learn to identify sources using training examples. Unless there is some definition of what a source is, performing source separation is not possible. This statement seems incongruous to our own experience as listeners. We grow up listening predominantly to mixtures, and we are able to identify sources even in the absence of spatial cues (e.g. multiple sources coming from the same speaker). In this paper we seek to design a system that exhibits the same behavior. To do so we aim to answer two questions: First, can we learn source models by only using mixtures with no ground truth source information? And second, can we use that training process in order to learn to perform single-channel source separation? What we find in this paper is that we can indeed learn to identify an open set of sound sources by training on spatial differences from multi-channel mixtures, and that we can transfer that knowledge to a system that can identify and extract sources from single-channel mixture recordings. We also find that this system works as well as had it been trained with clean data and accurate targets in a traditional manner.

To achieve our goal, we make use of the deep clustering framework in \cite{hershey2016deep}, because it exhibits the property of generalizing to multiple sources. This system learns embeddings of time-frequency bins of mixtures that are projected in distinct clusters corresponding to individual sources. This is ensured by using target labels that assign each input time-frequency bin to a cluster, and an appropriate cost function that finds an embedding that clusters the input similarly to the target labels. In our work, we will not make use of user-specified target labels, we will instead use an unsupervised approach where an appropriate embedding will be automatically discovered by forcing it to correlate with features extracted from the presented mixtures during training. Although there is a wide range of such features one can use, in this paper we will make use of inter-microphone phase differences as shown in \cite{rickard2007duet}. We additionally report results from using labels resulting from clustering these features, and leave it for future experimentation to explore more options. Once trained, this model can be applied on single-channel inputs, and as we show it can match, or even surpass, the performance of an equivalent model trained on user-provided labels. Through this approach we can minimize the effort required to construct training data sets, and instead learn such models by having them listen to multi-channel recordings in the field.
\section{Inferring Source Assignments}
\label{sec:Method}
In this section we show how we make use of spatial information to obtain targets that we can use to train a deep clustering network from mixtures. We will do so using two instances of a specific representation, but alternative features (not necessarily spatial) that also result in source clusters can be used instead. In the following sections we present how we constructed mixtures to test our proposed model, how we extracted the necessary features, and how we can infer a source separation mask.
\subsection{Input assumptions}
\label{sec:Method:Room Simulator}
\begin{figure}[h!]
\centering
\includegraphics[width=1.\linewidth]{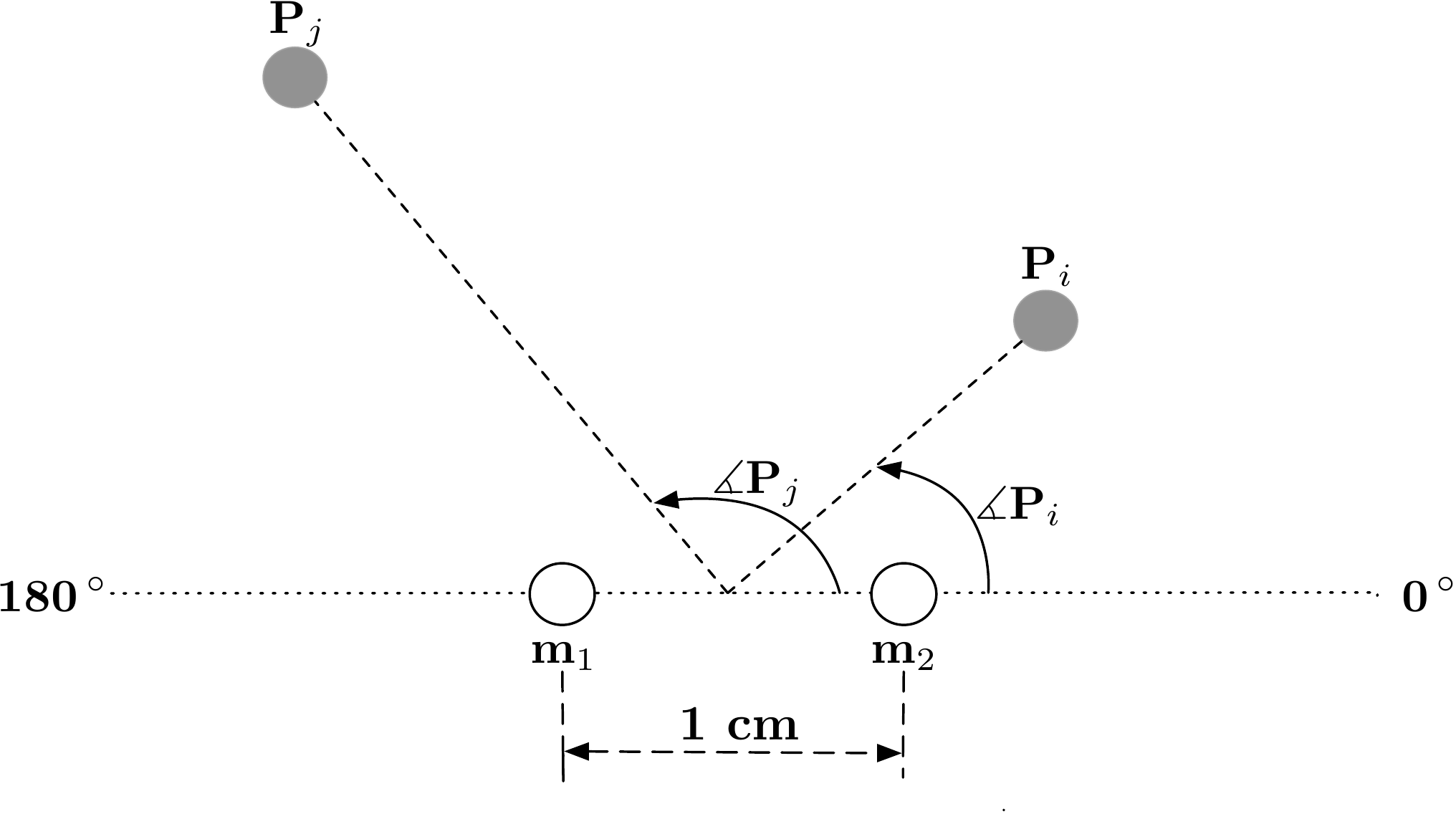}
\caption{Mixture setup for our experiments.}
\label{fig:RoomSimulator}
\end{figure}

We assume a mixture recording made with $2$ microphones that are set close together in the center of a room and $N$ sources in positions $\textbf{P}_1, \cdots, \textbf{P}_N$ which are dispersed across the area in fixed positions as shown in Figure \ref{fig:RoomSimulator}. We assume that the microphones are close enough that the maximal time-delay of a source between them will not be more than one sample. Furthermore, in order to have some spatial separation of the sources we assume that the angle difference between any two sources would be bigger than ten degrees, i.e., $|\angle(\textbf{P}_i) - \angle(\textbf{P}_j)|>10^o \enskip \forall \{i,j\}, i\neq j$. In our experiments we used a sample rate of $16$kHz, and a microphone distance of $1$cm.
Let $\textbf{s}_i(t)$ denote the signal which is emitted from source $i$ while a delayed version of it by $\tau>0$ would be noted as $\textbf{s}_i(t-\tau)$. Without any loss of generality, the mixtures that are recorded in both microphones ($\textbf{m}_1(t)$ and $\textbf{m}_2(t)$) can be written as:
\begin{equation}
\label{eq:mixtureintime}
\begin{gathered}
\textbf{m}_1(t) = a_1 \cdot \textbf{s}_1(t) + \cdots + a_N \cdot \textbf{s}_N(t) \\
\textbf{m}_2(t) = a_1 \cdot \textbf{s}_1(t+\delta \tau_1 ) + \cdots + a_N \cdot \textbf{s}_N(t+\delta \tau_N ) 
\end{gathered}
\end{equation}
where $\delta \tau_i$ denotes the time delay ($\delta \tau_i > 0$) between the recording of the signal $s_i$ from the first microphone to the second. The scalar weights $\{a_i\}_{i=1}^{N}$ reflect the contribution of each source in the mixture with the mild constraint of $\sum_{i=1}^{N} a_i = 1$ which makes our model very general. Thus, we can express the above equation in terms of the Short-Time Fourier Transform (STFT):
\begin{equation}
\label{eq:mixtureinfourier}
\begin{gathered}
\begin{bmatrix}
\textbf{M}_1(\omega, m) \\
\textbf{M}_2(\omega, m) 
\end{bmatrix} = 
\begin{bmatrix} 
1 \; \; \; \; \cdots \; \; \; \; \; 1  \\
e^{j \omega \delta \tau_1} \; \cdots \; e^{j \omega \delta \tau_N}
\end{bmatrix} \cdot
\begin{bmatrix}
a_1 \cdot \textbf{S}_1(\omega, m) \\
\vdots \\ 
a_N \cdot  \textbf{S}_N(\omega, m)
\end{bmatrix}
\end{gathered}
\end{equation}
where $\omega$ and $m$ denote the index of the frequency and time-window bins respectively of an STFT representation. 
\subsection{Phase Difference Features}
\label{sec:Method:Phase Difference Features}
\begin{figure}
     \centering
	\includegraphics[width=\linewidth]{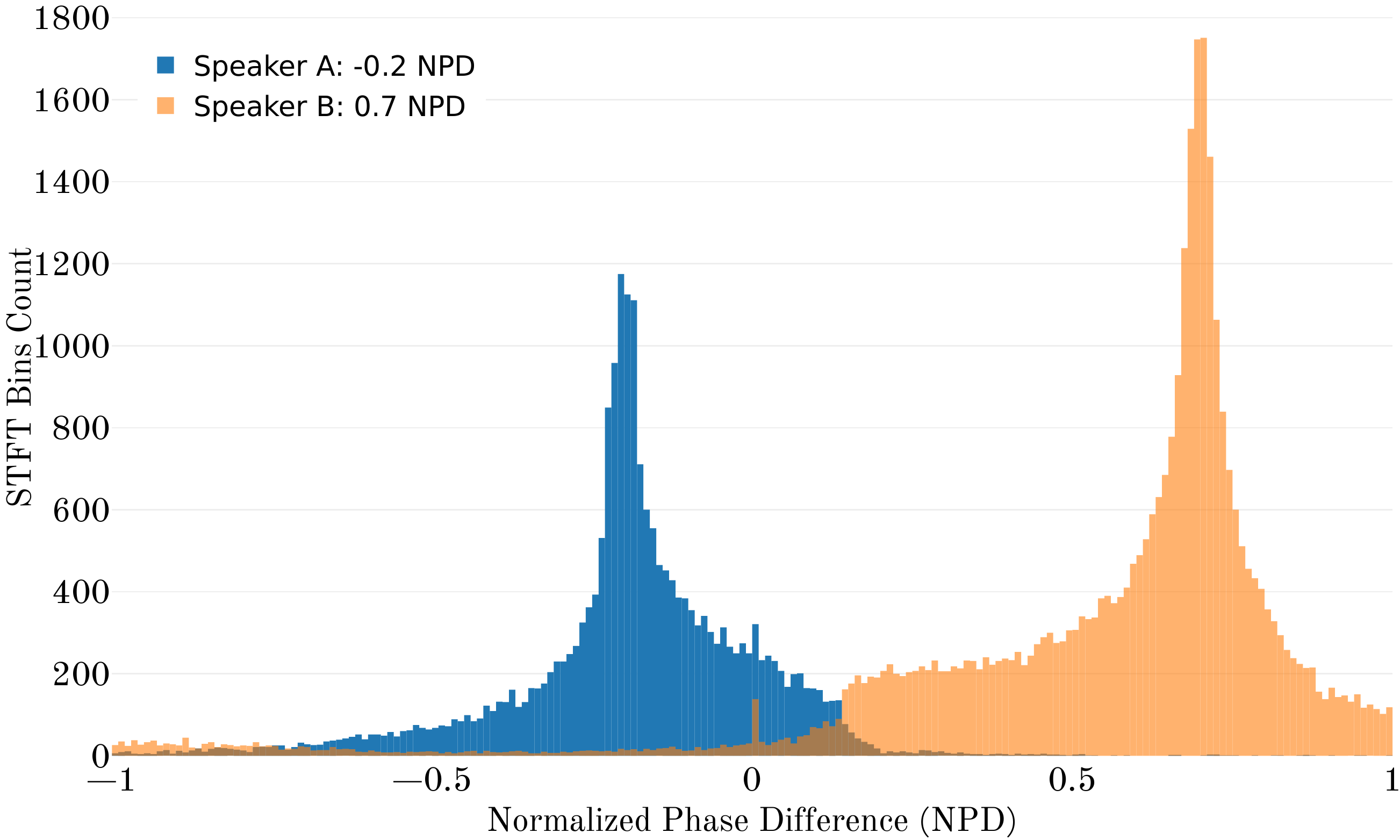}
    \caption{Histograms of normalized phase difference for a mixture of two speakers which are spatially separated. These features form clusters that reveal the two sources.}
    \label{fig:phasedifference}
\end{figure}

As long as $2 \pi F_s |\delta \tau_i| \le \pi$ and the sources are not active in the same time-frequency bins (W-joint orthogonality), the phase difference between the two microphone recordings can be used to identify the involved sources \cite{rickard2007duet}. Specifically, we consider a normalized phase difference between the two microphone recordings which is defined as follows: 
\begin{equation}
\label{eq:normalizedphasedifference}
\begin{gathered}
\bm{\delta \phi} (\omega, m) = \frac{1}{\omega} \angle \frac{\textbf{M}_1(\omega, m)}{\textbf{M}_2(\omega, m)}
\end{gathered}
\end{equation}
If the sources are separated spatially, we expect that the Normalized Phase Difference (NPD) feature would form $N$ clusters around a phase difference corresponding to each source's spatial position. Consequently, if the NPD of an STFT bin lies close to the $i$th cluster, we would expect that source $i$ dominates in this bin. As a result, the NPD form clusters that imply the sources across the STFT representation. We can also validate this statement empirically as shown  in Figure \ref{fig:phasedifference} where we plot the histograms of the NPD feature for all the bins of an STFT mixture with two speakers. The first speaker's NPD histogram (left) is well separated from the other speaker's (right).
\subsection{Source Separation Mask Inference}
\label{sec:Method:Separation Masks Inference}
Conventional supervised neural net approaches in speech source separation assume knowledge of the Dominant Source (DS) for each STFT bin of the input mixture \cite{huang2014deep} during training. Using that, one can define a target separation mask by assigning to each STFT bin a one-hot vector, where a $1$ is placed at the index of the source with the highest energy contribution. In our case, this can be written formally as:
\begin{equation}
\label{eq:DominatingSource}
\begin{gathered}
\hat{\mathbf{Y}}_{\mathit{DS}} (\omega, m, i) = \left\{
\begin{array}{ll}
      1 & i=\underset{1 \le j \le N}{\operatorname{argmax} } \left( a_j \cdot|\textbf{S}_j(\omega, m)| \right)  \\
      0 & \text{otherwise} 
\end{array} 
\right. 
\end{gathered}
\end{equation}
In our proposed model we wish to learn directly from mixtures and thus we use the NPD in order to infer a separation mask. We can do so by performing a K-means clustering on the extracted NPD features $\bm{\delta \phi (\omega, m)}$ where $K=N$ (assuming that we know the number of sources). Denoting the clustering assignment by $\mathcal{R}(\omega, m) = \{1, \cdots, N\} \; \forall \omega, m$ we express the mask we obtain from NPD features similarly to Eq. \ref{eq:DominatingSource} as: 
\begin{equation}
\label{eq:PhaseDifferenceBinaryMAsk}
\begin{gathered}
\hat{\mathbf{Y}}_{\mathit{BPD}} (\omega, m, i) = \left\{
\begin{array}{ll}
      1 & i= \mathcal{R}(\omega, m)  \\
      0 & \text{otherwise} 
\end{array} 
\right. 
\end{gathered}
\end{equation}
where the subscript BPD stands for Binary Phase Difference.
\section{Deep Clustering from Mixtures}
\label{sec:DeepClusteringProposed}
We can now propose direct learning from mixtures which is performed by giving as an input only the spectrogram of the first microphone recording, namely $|\textbf{M}_1(\omega, m)|$. This enables our model to be deployed and evaluated in situations where only one microphone is available. To simplify notation, we denote with $\textbf{X} \in \mathbb{R}^L$ the vectorized version of our input spectrogram $\textbf{X} = \operatorname{vec}(|\textbf{M}_1(\omega, m)|)$, where $L$ is equal to the number of STFT bins\footnote{Operator $\operatorname{vec}(\textbf{X})$ concatenates the columns of matrix $\textbf{X}$ into a one-column vector.}. Similarly, we express the vectorized versions of the masks in Eq. \ref{eq:DominatingSource}, \ref{eq:PhaseDifferenceBinaryMAsk} as $\textbf{Y}_{\mathit{DS}} = \operatorname{vec}(\hat{\textbf{Y}}_{\mathit{DS}})$ and $\textbf{Y}_{\mathit{BPD}} = \operatorname{vec}(\hat{\textbf{Y}}_{\mathit{BPD}})$, respectively. 
\subsection{Model Architecture}
Using the general structure in \cite{hershey2016deep}, our model encodes the temporal information of the spectrogram using stacked Bidirectional Long Short-Term Memory (BLSTM) \cite{hochreiter1997long} layers. To produce an embedding vector of dimensions $K$ we apply a dense layer on the output of the BLSTM encoder. The embedding output of our model can be seen as a matrix $\textbf{V}_{\bm{\theta}} \in \mathbb{R}^{L \times K}$, whose values should cluster according to the dominant source in each bin.
\subsection{Model Training}
\label{sec:directMixtureTRainingModelTraining}
An optimal clustering from such a network would be obtained if the encoded vector $\textbf{V}_{\bm{\theta}}$ would preserve the intrinsic geometry of the provided partitioning vector of the STFT bins for the $N$ sources, namely: $\textbf{Y} \in \mathbb{R}^{L \times N}$. For instance, in the ideal case that we know the dominating source, a partitioning vector for training would be obtained by Eq. \ref{eq:DominatingSource} (e.g. $\textbf{Y} = \textbf{Y}_{\mathit{DS}}$). In this context, we train our model using a weighted version of the Frobenius-norm loss function ($||\textbf{V}_{\bm{\theta}} \cdot \textbf{V}_{\bm{\theta}}^{\top} - \textbf{Y} \cdot \textbf{Y}^{\top}||_F^2$, presented in \cite{hershey2016deep}) that strives to push the self-similarity of the learned embeddings $\textbf{V}_{\bm{\theta}} \cdot \textbf{V}_{\bm{\theta}}^{\top}$ so as to best preserve the inner product of the provided partitioning vectors $\textbf{Y} \cdot \textbf{Y}^{\top}$:
\begin{equation}
\label{eq:Lossfunction}
\begin{gathered}
\mathcal{L}(\bm{\theta}) 
= \frac{1}{K} ||\textbf{V}_{\bm{\theta}}^{\top} \cdot \textbf{V}_{\bm{\theta}}||_F 
+ \frac{1}{N} ||\textbf{Y}^{\top} \cdot \textbf{Y}||_F \\
- \frac{2}{\sqrt{K N}} ||\textbf{V}_{\bm{\theta}}^{\top} \cdot \textbf{Y}||_F 
\end{gathered}
\end{equation}
where $||\textbf{A}||_F= \sqrt{\operatorname{trace}(\mathbf{A}\cdot\mathbf{A}^\top)} $ denotes the Frobenius norm of matrix $\textbf{A}$. This loss function can be efficiently implemented using Eq. \ref{eq:Lossfunction}, we avoid the computation of the large matrices $\textbf{V}_{\bm{\theta}} \cdot \textbf{V}_{\bm{\theta}}^{\top}$ and $\textbf{Y} \cdot \textbf{Y}^{\top}$. As shown in \cite{hershey2016deep}, this loss function pushes the model to learn an embedding space amenable to a K-means clustering that reveals the mixture sources. 
However, this loss function does not have to specifically use the partitioning vectors of the form $\textbf{Y} \in \mathbb{R}^{L \times N}$ and enables us to work with various target self-similarities $\textbf{Y} \cdot \textbf{Y}^{\top}$ obtained by any partitioning vector $\textbf{Y} \in \mathbb{R}^{L \times C}$, where $C$ is the number of features representing each STFT bin. In order to show the generalization that we are able to obtain by the proposed model, the following cases of $\textbf{Y}$ are considered: 
\noindent
\begin{itemize}
\item \textbf{Dominant Source (DS) mask}: $\textbf{Y}=\textbf{Y}_{\mathit{DS}} \in \{0,1\}^{L \times N}$ which is the one-hot mask obtained by assigning the dominant source on each STFT bin (Eq. \ref{eq:DominatingSource}). Obtaining this mask requires knowing how this mixture was designed.
\item \textbf{Unsupervised Binary Phase Difference (BPD) mask}: $\textbf{Y}=\textbf{Y}_{\mathit{BPD}} \in \{0,1\}^{L \times N}$ which is the one-hot mask obtained by performing K-means on the NPD features (Eq. \ref{eq:PhaseDifferenceBinaryMAsk}). This mask is automatically derived from features, and can be seen as implicitly separating the constituent sources based on spatial information and using that to train the network.
\item \textbf{Unsupervised Raw Phase Difference (RPD) features}: $\textbf{Y}=\operatorname{vec}(\bm{\delta \phi}(\omega, m))= \textbf{Y}_{\mathit{RPD}} \in \mathbb{R}^{L \times 1}$ which is the vectorized form of the raw NPD features (Eq. \ref{eq:normalizedphasedifference}). In this case, we do not explicitly estimate a separation mask. We instead compute raw spatial features which we should cluster according to the constituent sources. We thus train the network using a set of clusters (or embedding if you like), and not explicit source labels.
This latter model is the most important since instead of NPDs we can use any feature vector that is source-dependent and not worry about providing separation masks in any form. This allows us to think of the separation process as not just a regression on labels, but rather as finding a similar projection to a source-revealing embedding.
\end{itemize}
\subsection{Source Separation via Clustering on the Embedding Space}
\label{sec:deepClusteringProposed:Evaluation}
Assuming that the network is trained as described above with any variant of the label vectors $\textbf{Y}$, we now define how to perform source separation for a new input $\textbf{X}'$ using the output of our model $\textbf{V}_{\bm{\theta}}'$. To do so we perform K-means clustering on the embedding space with $K=N'$, where $N'$ is the number of different sources we expect to separate. As a result, a source separation mask is constructed for each input spectrogram $|\textbf{M}'(\omega, m)|$ based on the cluster assignments. After applying the predicted mask on the STFT of the input mixture $\textbf{M}'(\omega, m)$, we can reconstruct $N'$ signals on time domain $\hat{\textbf{s}}_1(t), \cdots, \hat{\textbf{s}}_{N'}(t)$. Note that in this case we do not need a multi-channel input, this operation is applied on a single-channel input.
\section{Experimental Setup}
\label{sec:Experiments}
\subsection{Generated Mixture Datasets}
For our experiments we used the TIMIT dataset \cite{garofolo1993timit} in order to generate training mixtures as described in Section \ref{sec:Method:Room Simulator}. For each mixture, the position of the sources is picked randomly leading to random time delays $\delta\tau_i$ and corresponding weights $a_i$ in Eq. \ref{eq:mixtureintime}. Each speech mixture has a duration of $2$s where all speakers are active. Six independent datasets are generated which are grouped by the number of active speakers $N=\{2, 3\}$ and their genders. Specifically for the latter case, each mixture dataset contains either 1) only females, 2) only males or 3) both genders, denoted by $f$, $m$ and $\mathit{fm}$ respectively. For each dataset, we generate $5400$ ($3$h), $900$ ($0.5$h) and $1800$ ($1$h) utterances for training, evaluation and testing respectively. None of these three sets share common speakers.
\subsection{Training Process}
All of our models are trained only on $N=2$ active speakers and for all the variants of gender mixing $f$, $m$ and $\mathit{fm}$. The spectrogram of the first microphone recording ($257$ frequencies $\times$ $250$ time frames) is provided as the input to our model. The deep clustering models we compare are trained under all three setups described in Section \ref{sec:directMixtureTRainingModelTraining}, where $\textbf{Y}$ is either $\textbf{Y}_{\mathit{DS}}$ or $\textbf{Y}_{\mathit{BPD}}$ or $\textbf{Y}_{\mathit{RPD}}$. Regarding the network topology, we varied the number of layers $\{2,3\}$, number of hidden BLSTM nodes per layer $\{2^{9},2^{10},2^{11},2^{12}\}$, the depth of the embedding dense layer $K=\{2^{4},2^{5},2^{6}\}$, dropout rate of the final BLSTM layer $[0.3,0.8]$ and the learning rate $[10^{-4},10^{-3}]$ (using Adam optimizer). The results shown below are from the best performing configuration in each case.
\subsection{Evaluation}
We test our models with mixtures having using two and three speakers but we only use models that are trained on mixtures of two speakers. In all cases, we perform source separation through clustering of the embedding space as described in Section \ref{sec:deepClusteringProposed:Evaluation}. In the case of three speakers, there is no additional fine-tuning of the pre-trained models on two speakers, instead K-means with $K=3$ is performed for the resulting embedding. In order to evaluate separation quality we report the Source to Distortion Ratio (SDR) \cite{vincent2006performance} for the reconstructed source signals as obtained from our models, and compare them with the performance we obtain when directly applying the binary masks $\textbf{Y}_{\mathit{DS}}$ and $\textbf{Y}_{\mathit{BPD}}$ which are the expected upper bound of performance. Note that our model produces its output by observing a single-channel input mixture, whereas the upper bounds shown are constructed using the labels $\textbf{Y}_{\mathit{DS}}$ and $\textbf{Y}_{\mathit{BPD}}$, which besides being optimal oracle masks, make use of spatial information from two channels. We additionally provide the SDR improvement over the initial SDR values of the input mixtures of $-0.01$dB and $-3.66$dB for two and three speakers, respectively.
\section{Results}
In Figure \ref{fig:sdrfm} we show the distribution of the absolute SDR performance for $\mathit{fm}$ mixtures containing two and three speakers. ``DC RPD", ``DC BPD" and ``DC DS" notate our proposed deep clustering model trained using $\textbf{Y}_{\mathit{RPD}}$, $\textbf{Y}_{\mathit{BPD}}$ and $\textbf{Y}_{\mathit{DS}}$, respectively. ``DS" and ``BPD" show the performance of the oracle binary masks $\textbf{Y}_{\mathit{DS}}$ and $\textbf{Y}_{\mathit{BPD}}$, respectively. ``Initial" denotes the input mixtures. We also provide the mean SDR improvement over the initial mixture for all gender combinations ($f$, $\mathit{fm}$, $m$) and for two and three speakers in Tables \ref{t:SDRimprovementsfor2} and \ref{t:SDRimprovementsfor3}. Notably, our results are also comparable with the baseline ``DC DS'' model \cite{hershey2016deep} trained using mixtures from Wall Street Journal (WSJ) corpus and ground-truth masks. The initial mixture SDR was measured to be $0.16$dB and $-2.95$dB for two and three speakers, correspondingly. 
\subsection{Unsupervised vs Supervised Deep Clustering}
\begin{figure}[h!]
    \centering
	\includegraphics[width=1\linewidth]{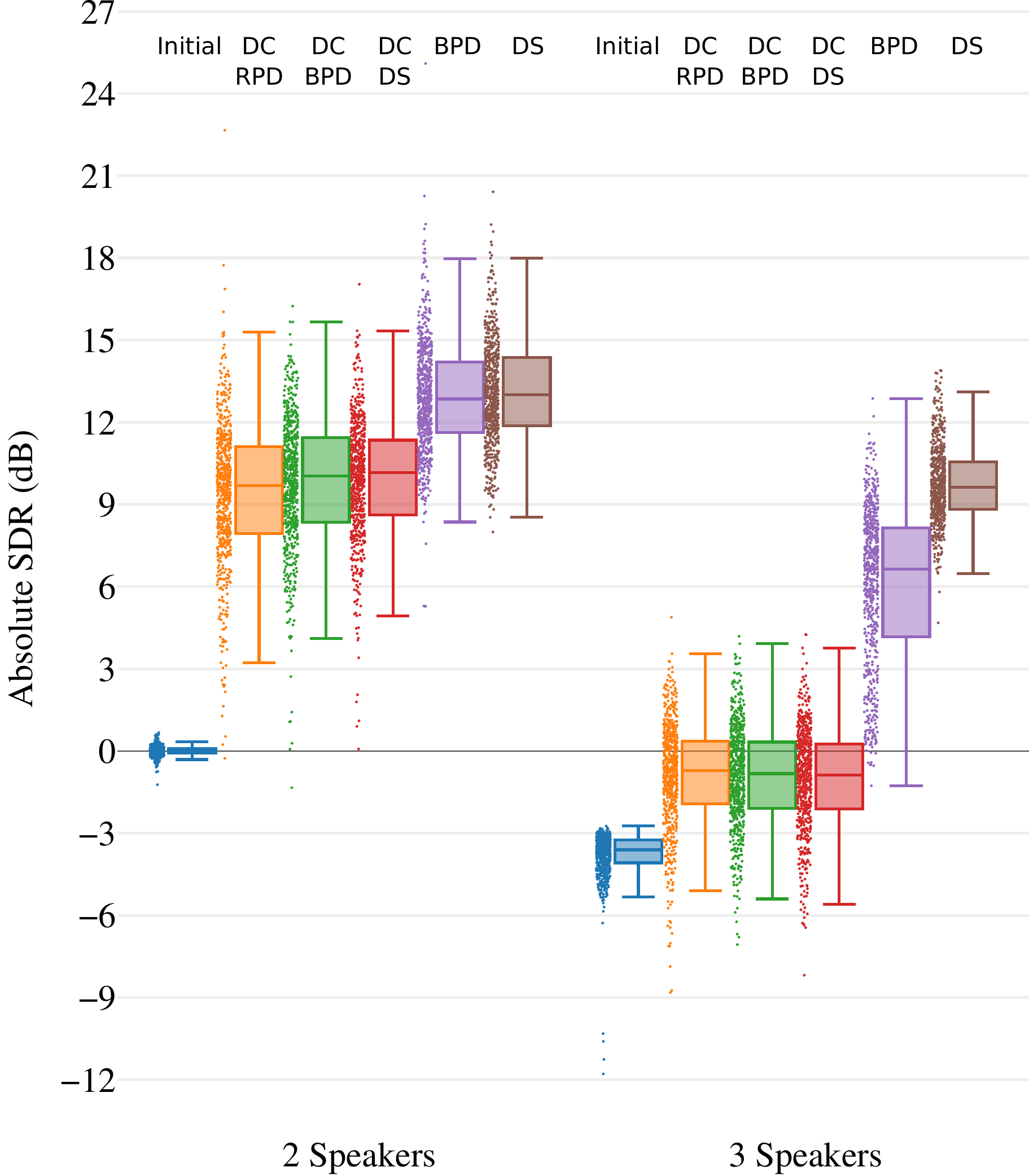}
    \caption{Absolute SDR performance for $fm$ mixtures containing two and three speakers. For each case we show the median and interquartile range (IQR) in the boxes, 1.5 times the IQR with the stems, and the raw data points as well. Note how training on mixtures and using targets based on spatial features results in equivalent performance to training on ground truth labels. Also, training on raw spatial features as opposed to a clustering defined from these features is roughly equivalent, which means that any spatial feature that forms clusters would work as well with our proposed model.}
    \label{fig:sdrfm}
\end{figure}

From Figure \ref{fig:sdrfm} it is clear that using the estimated source labels from $\textbf{Y}_{\mathit{BPD}}$ produce a similar upper bound to using $\textbf{Y}_{DS}$ which means that we provide adequate partitioning vectors for training our models without having access to the ground truth labels. This result is also reflected on the performance of our models. ``DC RPD" and ``DC BPD" which are trained in an unsupervised way from NPD features, perform just as well as the model ``DC DS" which was trained on ground truth labels. The median absolute SDR performance of the aforementioned models for two speakers of different genders are $9.73$dB, $10.16$dB and $10.28$dB, respectively, with a median upper bound of ``DS" ($13.14$dB) using ground truth and ``BPD" ($12.85$dB) using spatial information. SDR improvement obtained by our unsupervised models ``DC RPD'' and ``DC BPD" (Table \ref{t:SDRimprovementsfor2}) are also comparable with the baseline ``DC DS'' model \cite{hershey2016deep} with SDR improvement of $1.74$dB, $8.27$dB, $3.89$dB and $5.83$dB for the cases of $f$, $fm$, $m$ and $all$, respectively.
\begin{table}[h!]
\caption{SDR \underline{improvement} for two-speaker mixtures. The column $\mathit{all}$ denotes the average performance over same ($f$ and $m$) and different ($\mathit{fm}$) gender cases. Two last entries show the expected upper bounds for each target type.}
    \label{t:SDRimprovementsfor2}
	\centering
    \begin{tabular}{llcccc}
    &  &    $f$ &     $\mathit{fm}$ &      $m$ &      $\mathit{all}$  \\
    \hline 
    \hline 
    \multirow{3}{1.5cm}{Proposed} &
    DC RPD &   4.85 &   9.43 &   3.51 &   6.80 \\
    & DC BPD &   7.17 &   9.99 &   4.97 &   8.03 \\
    & DC DS  &   7.57 &  10.15 &   5.16 &   8.26 \\
\hline
    \multirow{2}{1.5cm}{Oracles} & BPD    &  13.65 &  12.88 &  11.82 &  12.81 \\
    & DS     &  14.02 &  13.19 &  12.14 &  13.14 \\
    \hline 
    \end{tabular}
\end{table}
\subsection{Generalization to Multiple Speakers}
When testing on three speakers, we notice a performance drop in the mean SDR improvement (see Table \ref{t:SDRimprovementsfor3}), which is consistent with previously reported results in deep clustering \cite{hershey2016deep} and \cite{isik2016single}. In this case, one should consider that our models were trained on two-speaker mixtures, and that when using three-speaker mixtures each source only accounts for a third of the mixture energy, thus being harder to separate. Regardless, we still see a net improvement on average over the initial mixture SDR for all gender cases ($1.75$dB for $f$ case, $2.75$dB for $fm$, $1.39$dB for $m$ and $2.16$ for $all$) when using our unsupervised model ``DC BPD". These results do not deviate significantly from the results obtained by ``DC DS" or the standard ``DC DS" \cite{hershey2016deep} supervised model ($2.22$dB).
\begin{table}[h!]
\caption{SDR \underline{improvement} in dB for three-speaker mixtures. The entries are similar to the ones in the table above.}
    \label{t:SDRimprovementsfor3}
	\centering
    \begin{tabular}{llcccc}
    \centering
     &   &   $f$ &     $\mathit{fm}$ &      $m$ &      $\mathit{all}$  \\
    \hline 
    \hline 
    \multirow{3}{1.5cm}{Proposed}  & DC RPD &   1.04 &   2.77 &   0.27 &   1.71 \\
     & DC BPD &   1.75 &   2.75 &   1.39 &   2.16 \\
     & DC DS  &   1.66 &   2.67 &   1.44 &   2.11 \\
    \hline 
    \multirow{2}{1.5cm}{Oracles} & BPD    &  10.23 &   9.76 &   8.83 &   9.64 \\
    & DS     &  13.88 &  13.44 &  12.41 &  13.29 \\
    \hline 
\end{tabular}
\end{table}
\section{Discussion and Conclusions}
\label{sec:conclusion}
In this paper, we present an approach that allows us to train on spatial mixtures in order to learn a neural network that can separate sources from monophonic mixtures. By using spatial statistics we allowed our system to learn source models without explicitly having to specify them. The advantage of this approach is that we do not need to provide ground truth labels, and that we let the system autonomously teach itself. We see that doing so can yield equivalent, or even better, results than by using a deep clustering model trained on ideal targets. Although we suggest a couple of different targets to use when training this system (namely the raw phase differences, or their cluster assignments), this method is amenable to using any other feature that would reveal source information (whether spatial or not). Also, the model we present here is a base deep clustering model proposed in \cite{hershey2016deep}. One can easily incorporate many of the latest deep clustering architectures and training processes (e.g. \cite{isik2016single}) to achieve better performance. Doing so is however out of the scope of this paper and is left for future work. Ultimately, we envision the approach we presented in this paper to allow us to design systems that can be deployed on the field and learn how to form source models in an unsupervised manner, as opposed to requiring training data designed by their users.
\bibliographystyle{IEEEbib}
\bibliography{mybib}
\end{document}